\documentclass{article}
\usepackage[utf8]{inputenc}
\usepackage{graphicx}
\usepackage{amssymb}
\usepackage{subfig}
\title{One-way Explainability Isn't The Message}
\author{Ashwin Srinvasan\thanks{AS is currently visiting the Centre of Health Informatics, Macquarie University and the School of CSE, UNSW.} \\
    \small Dept. of CSIS \& APPCAIR, \\
    \small BITS Pilani, Goa \\ \and
    Michael Bain \\
    \small School of CSE, \\
    \small University of NSW, Sydney \and
    Enrico Coiera \\
    \small Centre of Health Informatics, \\
    \small Macquarie University, Sydney
}
\date{\today}

\begin{document}

\maketitle

\begin{abstract}
Recent engineering developments in specialised
computational hardware, data-acquisition and storage technology
have seen the emergence of Machine Learning (ML) as a
powerful form of data analysis with widespread applicability
beyond its historical roots in the design of autonomous agents.
However---possibly because of its origins in the development
of agents capable of self-discovery---relatively little attention
has been paid to the interaction
between people and ML systems, although recent
developments on Explainable ML are expected to
address this, by providing visual and textual
feedback on how the ML system arrived at a conclusion.
In this paper we are concerned with the use of ML in
automated or semi-automated tools that assist one or more
human decision makers. We argue that requirements
on both human and machine
in this context are significantly different to
the use of ML either as part of autonomous agents for
self-discovery or as part statistical data analysis.
Our principal position is that the design of such human-machine
systems should be driven by repeated, two-way
{\em intelligibility\/} of information
rather than one-way explainability of the ML-system's
recommendations. Iterated rounds of intelligible
information exchange, we
think, will characterise the kinds of collaboration that will be needed to
understand complex phenomena for which neither man
or machine have complete answers.
To reassure the reader that this is not simply
wordplay, we propose operational principles--we
call them Intelligibility Axioms--to guide the design of
a first-step in constructing a
collaborative decision-support system. 
Specifically, for one iteration in
the collaboration (human-to-machine-to-human),
the principles are intended
to encode sufficient criteria for
the following:
(a) what it means for
information provided by the human to be intelligible
to the ML system; and
(b) what it means for an explanation provided
by an ML system to be intelligible to a human.
Using examples from the literature on the use of
ML for drug-design and in medicine,
we demonstrate cases where the conditions of the axioms
are met.
Intelligibility of communication is necessary, but not
sufficient to extend a single iteration of the collaborative loop
to multiple iterations. We
describe some additional requirements needed for the design of
a truly collaborative decision-support system.
\end{abstract}

\section{Introduction}
In the second half of his seminal 1950 paper~\cite{Turi:j:1950}, Alan Turing
describes an autonomous agent that has 
the capacity to alter its programming based on
experiments and mistakes, rather than relying on
human programmers. Since then, developments in
mathematics and computing have been making steady
progress in providing the groundwork for
Machine Learning, or ML.
But it is only recently that we have witnessed a sea-change in the
use of ML methods. Driving this change is a combination of the use of special-purpose
hardware, large amounts
of data from commodity devices and a significant decrease in the cost of
storing such data. This has meant that ML can now be part
of almost any kind of activity for which data can be collected and
analysed. In many such cases, the use of ML is
simply to predict accurately. Recently, one
form of ML -- deep neural networks -- have been able to achieve
startlingly good performance when provided with sufficient data
and sufficiently high computational resources. Progress is also
being made to replicate such success when the data are insufficient
\cite{transfer} and computational power is limited \cite{Kost:etal:p:2017}.
A difficulty has arisen, however, when attempting to
exploit the predictive performance of modern ML
techniques when the models they construct have to
be examined by humans who are not neural network specialists.
For example, a modern-day deep neural network may be able to
predict, with very high accuracy, the occurrence of malignancies
from X-ray images. If what is required is not just
what the prediction is, but also an explanation of how that prediction was arrived at, then
we hit an ``intelligibility bottleneck'', in making the explanation accessible to the
clinicians. Some of this arises from a mismatch between what certain ML practitioners view as suitable explanations, and what subject matter experts, or intended end-users, require~\cite{Amar:etal:x:2020}.
More generally, one could view this as a requirement for humans and ML systems to maximise their mutual knowledge, developed over sequences of communicative interactions~\cite{Coie:p:2001}.
Unfortunately some experts regard current techniques for explanation in ML as unfit for purpose, instead relegating ML systems to the status of drugs for which no definitive biological mechanism is understood, but which nonetheless may be effective in clinical practice, subject to the successful outcome of randomised controlled trials~\cite{Ghas:etal:j:2021}.
Whilst this may allow ML to be applied by clinicians in areas such as radiology, 
it falls short of ``human-level'' performance, at least as far as
explainability goes~\cite{Chan:Sieg:j:2019}.\footnote{
We recognise that there are clinical settings where explanations may {\em {not}\/} be required.
ARDA, for example, is a highly accurate ML-based tool for diagnosis of diabetic retinopathy (see: {\tt {https://health.google/caregivers/arda/}}).
based on the work reported in \cite{google:diabetes}. It has been trained using labelled
data provided by over 100 clinicians, and has been tested in a clinical trial. It is
a device for triage-assistance in settings where a clinician examines 1000s of patients
a day, and the tool is considered adequately field-tested. In this paper we are
concerned instead with what needs to be done if explanations {\em {are}\/} needed.}


But what does it mean for an explanation from an ML-system
to be understandable to the clinician, or,
more broadly, to a person who is a specialist,
but not in ML, interacting with a ML-based system?
At least some shared understanding of the concepts and terminology in a domain appears
to be needed for communication between human and machine, just as between humans.
Sadly, serious consequences may follow when this is lacking, for example, in the misinterpretation (wilfully or otherwise) of scientific knowledge in legal proceedings~\cite{Hodg:j:2000,Siff:c:2013}.
Perhaps we should aim for the the kind of explanations
that underlie interactions between humans with expertise in a specific area?
After all, much of scientific and medical progress has been the result of
interaction between groups of such people. However,
exactly how such specialists understand each other
rarely discussed.
Peter Medawar provides some clues:

\begin{quote}
Here then are some of the criteria used by scientists
when judging their colleagues' discoveries and the interpretations
put upon them. Foremost is their explanatory value -- their rank
in the grand hierarchy of explanations and their power to establish
new pedigrees of research and reasoning. A second is their clarifying
power, the degree to which they resolve what has hitherto been perplexing;
a third, the feat of originality involved in the research, the surprisingness
of the solution to which it led, and so on.	Scientists give weight (though
much less weight than mathematicians do) to the elegance of a solution.
({\em pg 52, Pluto's Republic\/})
\end{quote}

\noindent
It is unclear to what extent the design of
ML systems intended for interaction with people
explicitly take such criteria into
account. 
Even if they do so implicitly, certainly very little
is done to report them explicitly by way of explanation. 
At any rate, such intricate considerations may not be
needed when designing systems in which the role
of ML is as part of a tool for decision-support.\footnote{
Automated assistants for human specialists
are increasingly becoming necessary as the production and complexity
of data rapidly outpaces even the abilities of specialists to assimilate and process them. The distinction of ML-as-agent and ML-as-tool
in decision-making has been explored extensively
in \cite{federico}).} 
In this case, the ML-engine
is not expected to establish new pedigrees of research, or resolve
stubborn perplexities, and is also not the primary
decision-maker. 

In this substantially more restricted setting,
it seems entirely reasonable that the decision-maker
provides all relevant information at their
disposal to the tool, in a manner intelligible to
the tool. If nothing else, this may
avoid wasted effort on the part of the decision-support system
of re-discovering what is already known. It
would also seem to be pointless for the decision-support
tool to provide assistance in a manner unintelligible
to the person being assisted. These observations
lead us directly to a pair of operational criteria for checking
for two-way intelligibility between
a human specialist and an ML-based tool for decision
support.

\begin{figure}
    \centering
    \includegraphics[height=0.58\textwidth]{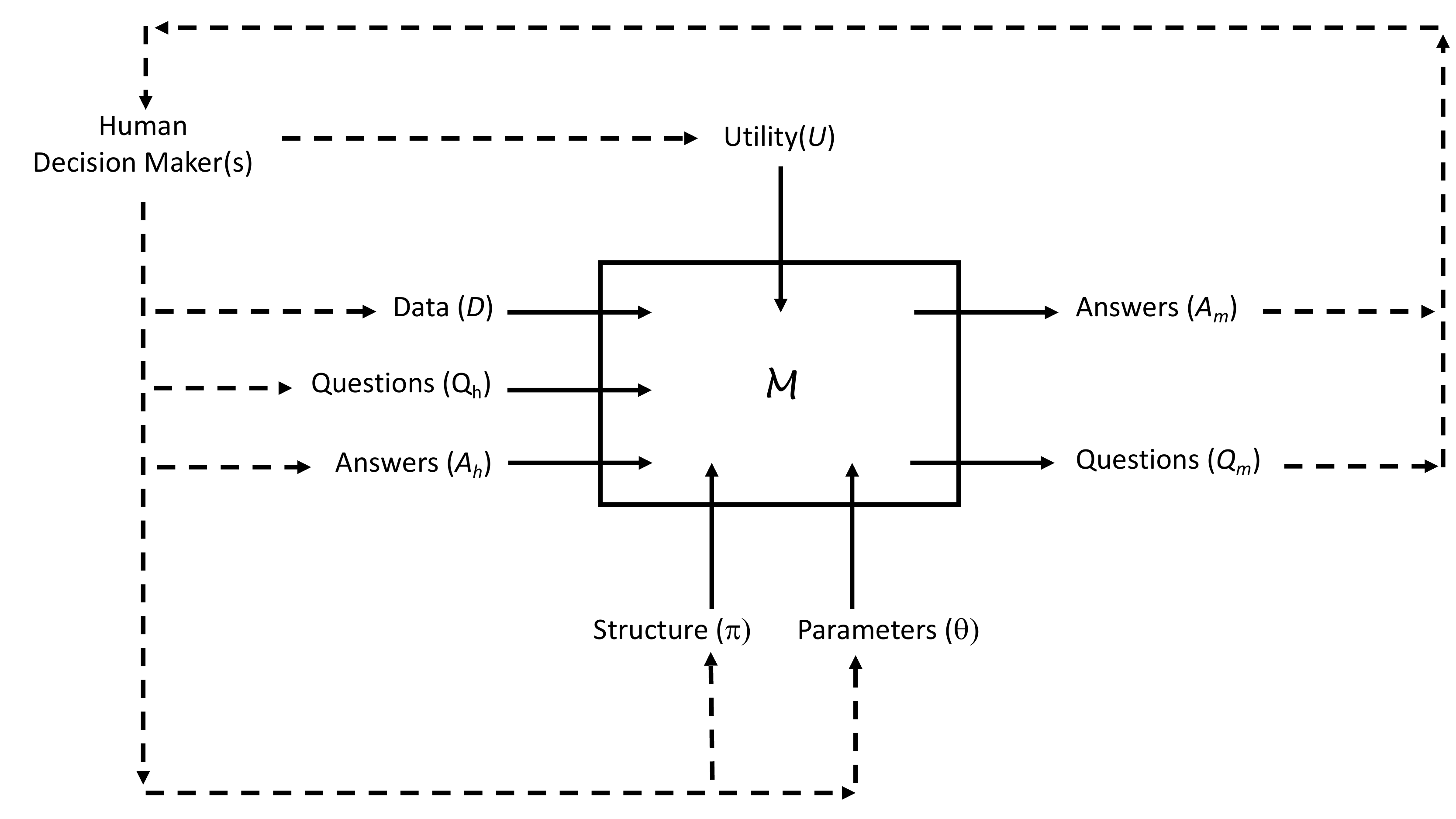}
    \caption{A generic ML system ${\cal M}$ for use
    in a decision-support system. The human decision-maker
    provides, at least: data $D$ and $U$, a utility function to be maximised.
    $\pi$ and $\theta$ are specifications of model-structures
    and parameters to be optimised: these may be specified by
    the decision-maker, or drawn by ${\cal M}$ from some pre-defined
    set. The performance of the machine will be obtained using $U$.
    The decision-maker's domain-knowledge can
    affect any of $D$, $U$, $\pi$ or $\theta$.
    Additionally, the decision-maker may also ask questions $Q_h$ of
    the machine
    (for example, ``what is the class-label
    of instance $x$?''), or provide answers $A_h$ to questions $Q_m$
    from the ML system
    which it
    can pose to the decision-maker (for
    example, a machine-generated question could be about the reason
    for a decision-maker's class-label for a data instance).
    The ML system provides answers $A_m$ to questions it receives.
    \emph{In this paper, explanations are taken to be a special
    kind of answer, provided by either human or machine.}}
    \label{fig:mlsys}
\end{figure}

Before presenting the axioms, we show in
Fig.~\ref{fig:mlsys} an abstract picture of the ML system
envisaged. For simplicity, we will refer to the ML system alone
as ``the machine'', even though there will be several other components in
a complete decision-support system. The machine
assists one or more human decision-maker(s), or domain-specialist(s), who may in turn use their
domain-knowledge to alter any or all
of $D$, $U$, $\pi$ or $\theta$ for the machine\footnote{
With reference to Fig.~\ref{fig:mlsys} it is unlikely that a human decision-maker with little
or no knowledge of the
inner workings of the ML system would nevertheless
be able to provide knowledge
to alter aspects like $\pi$ or $\theta$.
In practice, therefore, we would expect that
domain-knowledge would be used mainly to alter
$D$ or $U$. For the rest,
the ML-system may have to resort to some internal
mechanism for optimised selection from a set of pre-defined
alternatives. This set need not be finite, of course.}.
Again,
for simplicity, we will refer to the decision-maker(s), or domain-specialist(s),
as ``the human''.

\noindent
We propose intelligibility axioms in the following categories:

\begin{description}
    \item[Human-to-Machine.] This concerns
        intelligibility of the information provided
        by the human to the machine. For the present,
        we restrict the information to domain-knowledge and propose
        an axiom based on machine-performance:
            \begin{itemize}
                \item If the machine uses 
                    human domain-knowledge to improve its performance
                    then the domain-knowledge is
                    intelligible to the machine.
            \end{itemize}
    \item[Machine-to-Human Intelligibility.]
        This concerns the intelligibility to humans of
        information provided by the machine to account for
        a prediction. For the present, we will restrict
        this information to explanations, and propose
        an axiom on human refutability:
            \begin{itemize}
                \item If the human refutes the machine's
                    explanation
                    then the machine's explanation is intelligible
                    to the human.
            \end{itemize}
\end{description}
\noindent
 We will shortly provide examples from the literature
 where the premises of these axioms are satisfied. Later
 in the paper, we will also extend the axioms in each 
 category in order to characterise more fully the
 notions of human- and machine-intelligibility.
 
 At this point, the reader may be concerned about an
apparent circularity.
For example, in order for the machine to use the domain-knowledge,
doesn't it have to be intelligible in the first place? Similarly,
what does it mean to refute an unintelligible explanation? In fact, as conditional statements the
axioms identify human- or machine-intelligibility as necessary conditions for
some actions by machine or human (using, refuting),
but the actions themselves are only sufficient to infer
intelligibility. Thus, it is possible, for example, that the
human may not refute an explanation, and yet the explanation
may be intelligible to the human.

Informally, we will say ``human intelligibility holds'' to mean
the premise of the computer-performance axiom in
the human-to-machine category holds. Similarly, by
``machine intelligibility holds'', we will mean that
the premise of the human-refutability axiom is true.

We note that, in each category, intelligibility can be seen as a ternary
relation involving: the information-provider, the information provided,
and the information-recipient. In the literature on Explainable ML,
this has re-emerged as important requirement for
the acceptability of ML (see \cite{Mill:j:2019} for a recent example citing earlier work~\cite{Hilt:j:1990},
and~\cite{dm:ml} for an early identification of this).

The remainder of the paper is organised as follows.
In Sections~\ref{sec:humint} and~\ref{sec:machint}
we clarify each axiom in turn. In Section~\ref{sec:sysdesign},
we explore some consequences of taking the axioms into
account when designing  an ML-based decision-support tool. Section
\ref{sec:concl} concludes the paper.



\section{Two-way Intelligibility}

\subsection{Human-to-Machine Intelligibility}
\label{sec:humint}

\begin{figure}
    \centering
    \begin{tabular}{ccc} \\
    \includegraphics[height=3cm]{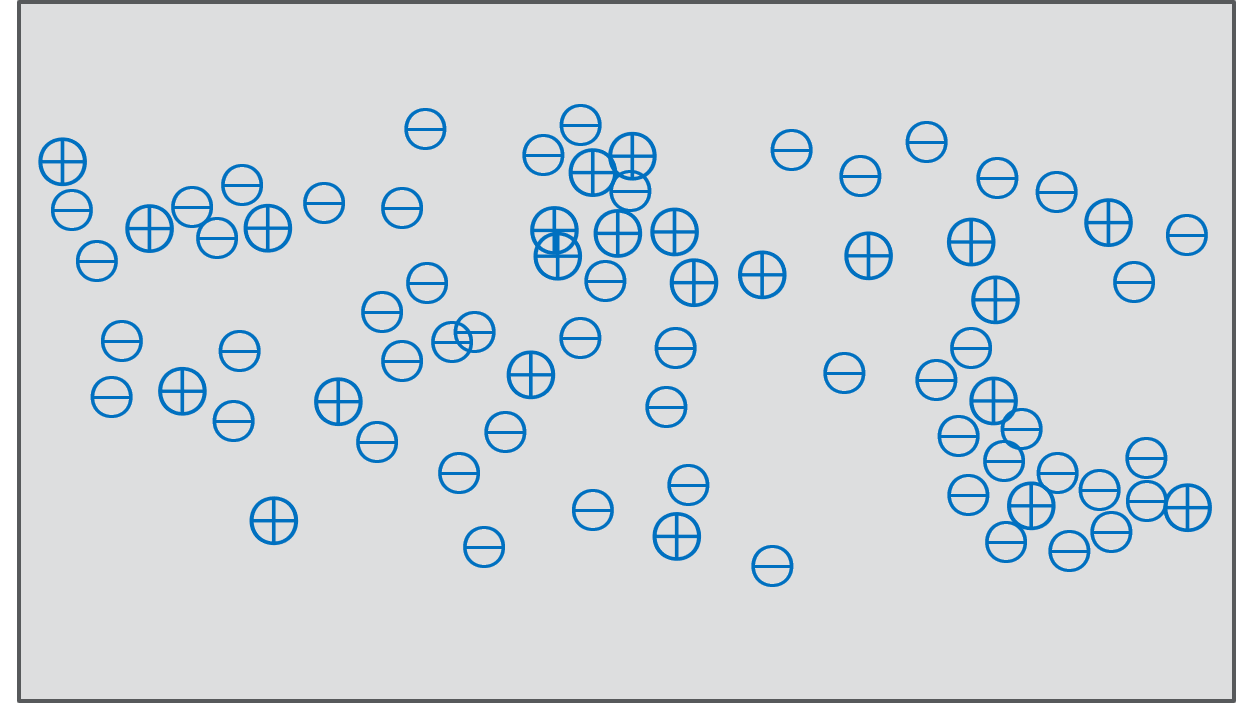} & \hspace*{1cm} & \includegraphics[height=3cm]{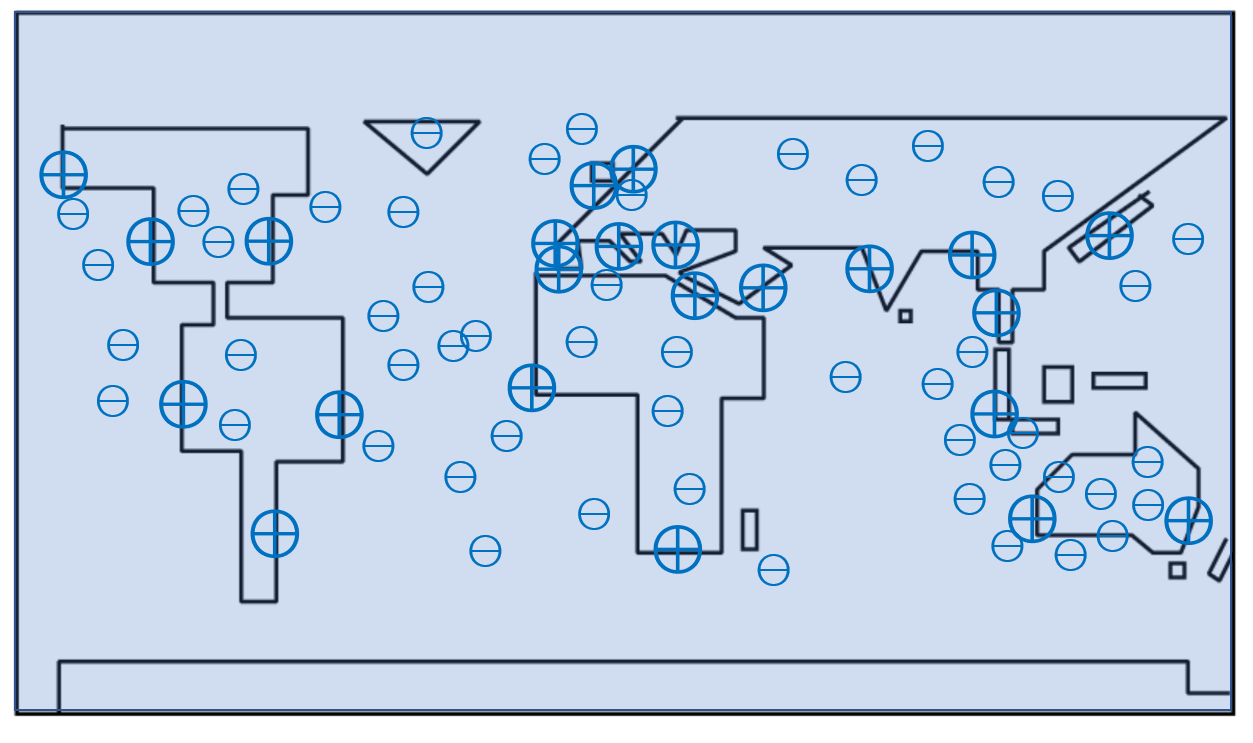} \\
    (a) & & (b) 
    \end{tabular}
    \caption{A classification problem: (a) Without information about the domain;
        and (b) With information about the points being on a world-map (from \cite{msb92}). The
        $\oplus$ points are all port cities, which can be easily inferred
        given knowledge of the land boundaries in (b).}
    \label{fig:world}
\end{figure}

We motivate the intelligibility to a machine of 
human domain-knowledge using an example
from \cite{msb92}, reproduced in Fig.~\ref{fig:world}.
Consider first the points in Fig.~\ref{fig:world}(a). Suppose
a machine was given the task of predicting accurately whether a new data
point was a $\oplus$ or a $\ominus$ from this data. The problem
is difficult, and can be made increasingly harder, to the
point of apparent randomness, with more data. Now the machine
is told -- let us say by a geographer -- what else is known (Fig.~\ref{fig:world}(b)). All at
once, predictive performance jumps from random-guessing to 100\%
accuracy simply from the machine inferring that the $\oplus$ points
are port cities. By the Human-to-Machine Intelligibility Axiom, the information
provided by the geographer is said to be intelligible to the machine.

Predictive accuracy is used here purely as an
example of a utility function that the human decision-maker (here, our
 geographer) wants the machine to maximise. At this point,
 it is useful to clarify that an increase in predictive performance
 is only postulated as being sufficient for claiming intelligibility.
 Nothing is said on about intelligibility if predictive
performance does not increase. For example, the machine
may already have an encoding of the information
provided by the map. 

Despite this example, it is still reasonable to ask why it
is necessary at all for the human decision-maker to provide
domain-knowledge to the ML system.
In principle, given  sufficient data and computational
resources, an ML engine--like a deep neural network--may be able to
reconstruct the domain-knowledge for itself.
In practice, the difficulty
lies in the qualification of ``sufficient data and computational resources''.
Neither may be available, and the ML engine can under-perform. Now the
human decision-maker either does or does not know that this limitation exists.
If the limitation is known,
then it seems perverse for the decision-maker
to withhold information that would help the machine do better,
or by providing it in a manner that the machine cannot use.
Alternatively, if the limitation is not known,
then the only reasonable course for the decision-maker
would appear to be to provide what information they have in a manner
the machine can use (and possibly ignore if it is redundant).
In either case it would appear best to provide the information
that is known in some machine-readable form.

\subsubsection*{Examples from Drug-Design}

The design of new drugs is well-known to be time-consuming and expensive process,
requiring the involvement of significant amounts of human chemical
expertise. It is also one where data can now be generated and tested automatically on
a very large scale. One phase in the pipeline of drug-design is that of early stage
discovery of ``leads''. These are small molecules that could potentially form the
basis of a new drug. ML-based models have been used to assist in this by constructing
models relating chemical structure to molecular activity (like binding affinity, toxicity,
solubility and so on), and even in the generation of entire new small molecules. Reports,
both new and old, in the literature have shown that provision of domain-knowledge can make
a significant difference to the performance of the ML engine involved.

We show first recent results reported in \cite{dash:botgnn,dash:drm}. The experiments
reported consider the inclusion of human-selected domain-knowledge for two kinds
of deep neural networks (a multi-layer perceptron, or MLP, and a graph neural network, or GNN).
The domain-knowledge consists of the definitions of approximately 100 relations encoding
functional groups and ring-structures. Fig.~\ref{fig:tox1} tabulates
the number of datasets on which performance improvements are observed,
with the inclusion of human domain-knowledge.
For each deep network type, for problems in the ``Better'' column,
the machine uses the information provided to improve performance.
In each such instance, the Human Intelligibility axiom will
infer that the domain-knowledge is intelligible to the machine, but say nothing about the remainder.

\begin{figure}
    \begin{center}
        \begin{tabular}{|l|ccc|} \hline
               & \multicolumn{3}{|c|}{Comparative Performance} \\
    DNN    & \multicolumn{3}{|c|}{(with domain-knowledge)} \\ \cline{2-4} 
        Type   & Better & Same & Worse \\ \hline
        MLP    &   71   &  0   &  2       \\
        GNN    &   63   &  9   &  1     \\ \hline 
        \end{tabular}
    \end{center}
    \caption{The use of domain-knowledge by two kinds
        of deep neural networks (DNNs): mult-layer perceptrons (MLPs) and graph-neural
        networks (GNNs). Estimates of performance are obtained on 73 different
        datasets. Here, ``Better'' (respectively,
        ``Same'' and ``Worse'') means
        the use of domain-knowledge results in an
        improvement in performance (respectively,
        no change, and worse performance).}
    \label{fig:tox1}
\end{figure}


Simply providing domain-knowledge does not guarantee intelligibility, as is
apparent from Fig.~\ref{fig:tox1}. A separate but useful point to note is that intelligibility
depends on the recipient of the information. That is, whether
or not the domain-knowledge provided is intelligible can depend
on the ML system used. This is evident in results reported in  
an early work reported in \cite{trep:back}. There
the same domain-knowledge is provided in the same representation
to two symbolic ML systems (say $A$ and $B$). The study examines the
progressive increase in domain-knowledge for a well-studied toxicology problem.
There a progressive increase in domain-knowledge does not affect
the performance of systems $A$ and $B$ in the same way (Fig.~\ref{fig:tox2}).

\begin{figure}
    \centering
    \begin{tabular}{|c|cc|} \hline
    Domain &  \multicolumn{2}{|c|}{Predictive Performance} \\
    Knowledge & System $A$ & System $B$ \\ \hline
    B1  & Increases & Increases \\
    B2  & Increases & Decreases \\
    B3  & No change & Increases \\
    B4  & Increases & No change \\ \hline
    \end{tabular}
    \caption{Comparison of the performance of 2 ML systems with subsets
        of domain-knowledge (based on results reported in \cite{trep:back}).
        Each subset contains the definitions from the previous subset
        (that is $B1 \subset B2 \subset B3 \subset B4)$. The initial
        change in predictive performance is measured against the
        baseline performance obtained without any domain-knowledge.
        Subsequent changes are measured against the performance
        obtained with the set just earlier in the sequence.
        Increase and decrease of performance
        are statistical assessments of improvements in
        predictive accuracy (that is, the
        increase in predictive accuracy has to be statistically significant).}
    \label{fig:tox2}
\end{figure}

Thus, with some subset of domain-knowledge ($B2$, for example) 
our axiom could conclude
it was intelligible to system $A$, but remain silent about system $B$,
or {\em vice versa\/} ($B3$, for example).\footnote{
Actually, most ML systems have several parameters. As
ML practitioners are well-aware, changing values of
parameters can greatly change the performance of the system.
Thus, intelligibility can vary even for a single ML system.
}

\subsection{Machine-to-Human Intelligibility}
\label{sec:machint}

The principal motivation for the intelligibility of a
machine's explanation is provided
by a description by Michie~\cite{Mich:John:b:1984} on the need or
otherwise for explainability of
machine-constructed answers, in the context of human decision-making.
We reproduce here some parts of the article that are relevant.
The description begins with the construction of a black box for a chess
endgame (the interjections in brackets are ours):

\begin{quote}
At the meeting in Toronto in 1977 of the International Federation for Information Processing, Kenneth Thompson of Bell Telephone Laboratories presented a computer program for playing the chess end-game of King and Queen against King and Rook. He had done this by the ultimate in `hammer and tongs' methods: in the absence of a complete set of rules for playing the end- game, he had previously programmed the machine to work out what to do in every single possible position \ldots All these moves were then loaded into a gigantic `look-up' table in the machine's memory \ldots
Thompson invited [International Masters] to demonstrate winning play for the Queen's side against the machine. To their embarrassment they found they could not win, even after many attempts  \ldots
The machine repeatedly conducted the defence in ways which to them were so bizarre
and counter-intuitive [like separating King and Rook] that they [the Chess Masters] were left grasping air \ldots
Naturally [they] found the experience upsetting. They wanted to ask the program to explain its strategy, but this of course neither it nor its author could do. The answer in every case was, `It's in the table.'
(``The strange case of Thompson's table'', pg. 64, {\em The Creative Computer\/})
\end{quote}

Michie describes how this situation is not very different to the case of
machine-learning programs that are unable to explain their decision-making.
He sees this as not being especially problematic in some circumstances.\footnote{
Surprisingly, he includes the possibility that scientific discovery
may even benefit from highly predictive but opaque machine-constructed
models, since it would force scientists to develop new explanations for
unexpected predictions.}
However in some other cases, involving decision-making in critical areas, the
lack of meaningful feedback from the machine can become a serious issue:

\begin{quote}
 But what if the system were doing something of social importance, such as managing a complex control function in factory automation, transport or defence? Two supervisors, let us imagine, are responsible for intervening manually in the event of malfunction. The system now does the equivalent in industrial or military terms of ‘separating its King and Rook’. ‘Is this a system malfunction?’ the supervisors ask each other. They turn to the system for enlightenment. But it simply returns the same answer over and over again \ldots

The problem becomes of global importance when the system being operated is in air traffic control, air defence or nuclear power. As control devices and their programs proliferate, their computations may more and more resemble magical mystery tours \ldots
Any socially responsible design for a system must make sure that its decisions are not only scrutable but refutable \ldots (``The lunatic black box'', pg. 68,
{\em The Creative Computer\/})
\end{quote}

\noindent
In this paper, we require that that a machine's decision
is always accompanied by an explanation. Thus, refutation of the decision
as described by Michie is taken to be tantamount to refution of the
explanation that goes with it. But here we arrive at a difficulty:
what exactly is meant by a refutation of an explanation? The term
has a well-understood
meaning in the natural and mathematical sciences. In the former, explanations are
hypotheses about natural phenomena and refutations follow from the result of experiments
devised to test assumptions and predictions. In the latter, explanations are proofs, which
can be refuted by demonstrating inconsistencies. In our case, we will treat models constructed
by an ML system as a hypothesis about the data, and an explanation will refer to
descriptions justifying the answer to a question posed to the machine. We will minimally
require that the refutation of an explanation will result in the explanation
being categorised by the decision-maker as being incomplete (insufficient) or incorrect.
As before, refutability of explanations will only constitute a sufficient criterion for intelligibility
of the machine's explanation to the human decision-maker.

\subsubsection*{Examples from Medical Informatics}

We illustrate the Machine Intelligibility Axiom using results obtained
in medical informatics. This area is chosen here for two reasons. First,
clinicians increasingly have access to data of many different kinds collected for individual patients, ranging
from traditional test-reports  to genomic information in the form of patient-specific single-nucleotide polymorphisms (SNPs).
Also available, with some additional effort, are results of treatments and outcomes from across
the world, and evidence and data from population studies. If clinical decision-making is to
deal effectively with the data, then some form of automated assistance seems inevitable.
Secondly, despite increased automation, decision-making is still expected to rest firmly
with the clinician, in all but routine monitoring systems. So, what we can expect
to see is an increased usage of automated decision-support tools.

\begin{figure}
    \begin{minipage}{\textwidth}
     \centerline{\includegraphics[height=8cm]{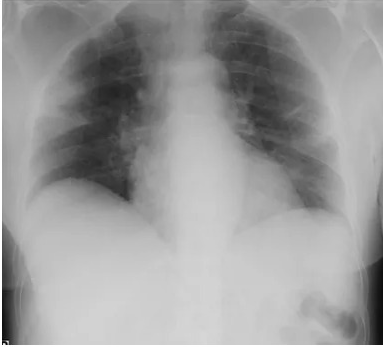}}
    \begin{center}
    {\scriptsize{\em
    \begin{tabular}{lll} \\
    Not Covid because: & \hspace*{0.2cm} &  The explanation does not mention \\
    \hspace*{0.1cm} Air-space opacification probability is low,  and & & the right upper lobe air space \\
    \hspace*{0.1cm} Cardiomegaly probability is high; and & & opacification consistent
            with Covid\\
    \hspace{0.1cm} Emphysema probability is low & & \\
    \hspace*{0.1cm} Pneumothorax probability is low \\
    \hspace*{0.1cm} Fibrosis probability is low \\[6pt]
    {\bf {Machine's explanation}} & & {\bf {Radiologist's feedback}}
    \end{tabular}
    }}
    \end{center}
    \end{minipage}
        \begin{minipage}{\textwidth}
        \vspace*{0.5cm}
        \begin{center}
            {\scriptsize{
                \begin{tabular}{|l|c|c|c|} \hline
                \multicolumn{4}{|c|}{Radiologist's Opinion about the Model's} \\ \hline
                Explanation & \multicolumn{3}{|c|}{Prediction} \\ \cline{2-4}
                & Correct   & Wrong & Unsure       \\ \hline
                Sufficient & 17 & 0  & 0\\
                Incomplete & 1 & 3 & 1\\
                Incorrect & 3 & 2 & 3\\ \hline
            \end{tabular}
            }}
        \end{center}

        \end{minipage}
    
    \caption{Top: A chest X-ray (the original image is of low quality);
        Middle: The machine's explanation and a senior radiologist's feedback; and
        Bottom: A tabulation of the
        radiologist's assessment of explanations from the ML-based system
        on a set of test images.}
    \label{fig:covidx}
\end{figure}

First, we look at a recent research study on identification of
Covid-19 patients, based on X-ray images. The automated tool described in
\cite{covid} uses a hierarchical design in which clinically relevant features
are extracted from X-ray images using state-of-the-art deep neural networks.
Deep neural networks are used to extract
features like ground-glass opacity from the X-rays; and the system
also includes a network for
a deep network for prediction of possible disease (like pneumonia).
The output from
the deep networks are used by a symbolic decision-tree learner to
arrive at a prediction about Covid-19. Explanations are textual
descriptions obtained from the path followed by the decision-tree.
Results reported in \cite{covid}
describe how this neural-symbolic approach compares to an end-to-end monolithic
neural approach (the predictive results of the two are comparable). However, our
interest here is on the clinical assessment of the explanations produced
by the symbolic model by radiologists: Fig.~\ref{fig:covidx} shows
an example of a machine's explanation and a clinician's assessment
of that explanation. A tabulation of assessment on several ``test'' images
is also shown.  From the tabulation we can see:
(a) The radiologist
does not always think the model is correct (this is despite a
supposed predictive accuracy of over 99\% claimed for the model);
(b) The radiologist is more likely to refute the explanation when he
thinks the model is wrong;
(c) Overall, the radiologist refutes
the explanations in a substantial proportion of instances (13/30 $\approx$ 43\% of the
instances). For us, it is (c) that is most relevant. With the Machine Intelligibility
Axiom, the machine's explanation is concluded as being intelligible for the 13 instances for which
are refuted; and no inference is made on the 17 instances where
the radiologist has rated the explanation as sufficient.

To the best of our knowledge,  the most direct example
of refutations in medical informatics is from an early decision-support
tool in chemical pathology.
PEIRS \cite{PEIRS:j:1993} was an extremely successful
decision-support tool for a pathologist that constantly revised
a tree-structured model consisting of rules and exceptions.
PEIRS relied entirely on the ability of the pathologist
to read, understand and refute the model's explanation. The
refutation in turn triggers a revision to the machine's model.
But is a 30-year old tool even relevant, given today's ML? We think
the answer is ``yes'': decision-support tools using the PEIRS approach
have continued to be deployed and used with great success \cite{Comp:Kang:b:2021}.
Much of this is based directly on the approach pioneered by PEIRS. In turn, they
inherit some of the ML-limitations of PEIRS: we will return to these later in
the paper.

In both applications we have described there are instances where
the human does not refute the machine's explanation. Again,
that this does not imply unintelligibility. But it does suggest
there is room to expand the concept of machine-intelligibility
to include some (but not all) instances when explanations are not refuted.
We will propose such a modification later in the paper.
Finally, as with the previous axiom, intelligibility continues to be
recipient dependent --- a repeat experiment with a second radiologist found
that the explanations refuted were different. Those results are not shown
here.

\subsection{Two Additional Axioms}

There is a deficiency in the characterisation of
intelligibility we have presented so far. The axioms have
nothing to say about when
a human gains anything from the machine's explanation. Also, there
is nothing in the axioms about when a human's explanation being intelligible
to a machine. In a truly useful human-machine collaboration, we would expect
both to be relevant. We therefore extend the list of axioms to be
more balanced:
    
    \begin{description}
        \item[Human-to-Machine (contd.)] We propose
            a ``machine-refutability'' axiom:
            \begin{itemize}
                \item If a machine refutes a human's explanation
                    then the human's explanation is intelligible to the machine
            \end{itemize}
        \item[Machine-to-Human (contd.)] We propose a
            ``human-performance'' axiom:
    \begin{itemize}
            \item If the human uses the machine's explanation to
                improve performance
            then the machine's explanation is intelligible to the human\footnote{
                We note that this axiom will hold with the use of
                what Michie calls ``Ultra-Strong'' ML \cite{dm:ml}.}.
    \end{itemize}
\end{description}

Although the human-performance axiom mirrors the computer-performance
axiom presented earlier,  more care will be needed in practice with
 an ``open system'' like a human to ensure that
improved performance is indeed connected to the use of the machine's explanation
and not due to some other factor (see \cite{mugg:expl} for
an experimental design that attempts to identify conditions under which
machine-generated explanations can be directly attributable to improvements
in performance).

\section{Designing for Two-Way Intelligibility}
\label{sec:sysdesign}

Thus far we have argued that to enable useful collaboration between
human and machine information from one must be intelligible to
the other; and we have identified actions by the human and machine
that allow us to infer intelligibility. But several aspects remain unspecified.
What constitutes an explanation? How does a human or
a machine refute an explanation? Can refutations be themselves be
refuted? And so on. These are questions that arise in the design of a 
decision-support system with ``built-in'' intelligibility. 
We are as yet in too early a stage in the design of such systems
to provide normative definitions for these concepts. Instead, we attempt to
understand them better by focusing
on the restricted task of learning to classify observations.

\subsection{Collaborative Classification}
\label{sec:collab}

First we examine some entities and relations
that characterise the interaction on the
$i^{\mathrm{th}}$ iteration of the collaboration ($i=0$ denotes
the initial condition, before any exchange of
information between human and machine):

\begin{enumerate}
    \item $X$, a set of instances to be classified and $L$ a
        set of labels;
    \item $T$, a function that
        classifies correctly all
        instances in $X$. $T$ will be called the {\em oracle\/}.
        $T$ is not known to the human or to the machine;
     \item $B_{h}^{(i)}$, the domain-knowledge of the human decision-maker,
        and $B_{m}^{(i)}$, the domain-knowledge of the machine, on the $i^{\mathrm{th}}$ iteration;
    \item $H_{B_h}^{(i)}$ is the human decision-maker's approximation
        to $T$, and $M_{B_m}^{(i)}$ is the machine's approximation to $T$.
        $H_{B_h}^{(i)}$ and $M_{B_h}^{(i)}$ will be called ``hypotheses''.
        We will assume $B_h^{(i)}$ is contained in $H_{B_h}^{(i)}$ and denote
        the latter by $H_i$. Similarly
        for $M_i$.
        In any iteration $i$, none of the $M$'s up to
        $i$ may be known completely
        to the human and none of the $H$'s up to
        $i$ may be known to the machine;
    \item The annotation of an instance $x_{H_i}$ using the hypothesis
    $H_i$ is the pair $(x,H_i(x))$. Similarly for
    $x_{M_i}$;
    \item $Send(X,\mu,Y)$ is a relation denoting that
        $X$ sends a message $\mu$ to $Y$. Here $X$ and
        $Y$ can be the human ($h$), machine ($m$) or
        oracle ($o$). $X$ and $Y$ are assumed to be
        distinct;
    \item $Receive(X,\mu,Y)$ denotes that $X$ receives
        a message $\mu$ from $Y$. As with $Send$, $X$
        and $Y$ are distinct, and can be one of $h$, $m$,
        or $o$; and
    \item $P_{H_i}$ denotes some estimate of the performance
        of the human, and $P_{M_i}$ the corresponding performance
        of the machine.
\end{enumerate}

\subsubsection{Messages}
\label{sec:message}

All collaboration is achieved through messages, which amounts to
a kind of conversation between human, machine and, to some extent,
with the oracle. Let us
assume for the moment that communication between human and
machine is instantaneous, noise-free, and free. But none of
these are true of communicating to the oracle, which may
take time, and is expensive. For the present also we will assume
that only the human communicates with the oracle, but will make
available to the machine information received from the oracle.
Of particular interest to us are messages of the following
kind:

\begin{description}
    \item[Questions.] Questions are often, though not always, expected
        to be about the annotation of an instance. We will distinguish messages with
        questions by the use of a ``?''.
        For example, $Send(h,(x,L)?,m)$ denotes a question posed by the
        human to the machine about the machine's class-label
        for instance $x$. Here $L$
        for variable whose value is to be returned
        by the machine). Once the message is received by
        the machine, 
        $Receive(m,(x,L)?,h)$ would be true. Questions to
        the oracle from the human are similarly denoted by
        $Send(h,(x,L)?,o)$.
    \item[Explanations.]  Explanations are a class of answers to
        questions. Messages in this class are of the form
        $A~because~R$, where $A$ denotes and answer and $R$ denotes
        the reason for the answer. Answers to questions about the annotation
        of an instance,
        for example, may be explanations of the form 
        $(x,l) ~because~\mathit{Proof}$
        where $\mathit{Proof}$ denotes some demonstration of
        the reasoning why an instance $x$ has the label $l$.\footnote{
            A proof for an annotated instance
            $(x,l)$ in a logical sense is specified easily enough
            as a function $\mathit{Proof}((x,l))$ that returns a 
            sequence of inference steps $\langle S_1,S_2,\ldots,S_k\rangle$.
            However, we do not commit to this form of proof-specification
            here.} We will use $\blacksquare$ to denote a special
            kind of reason for an answer $A$, which is to be read
            as ``$A$ is true''.
            Explanations from the oracle will always be of
            the form $A~because~\blacksquare$.
    \item[Refutations.] Refutations are messages that
            rebut explanations. We restrict
            refutations to explanations for which
            the reason isn't $\blacksquare$ (the oracle's explanation
            is therefore irrefutable). For an explanation $E$
            of the kind $(x,l)~because~p$, we distinguish
            two kinds of refutations based on (apparent) errors in the proof $p$:
        \begin{itemize}
            \item The proof $p$ is overly-specific.
                A possible refutation is then
                $(\mathit{overspecific}(E)$ $because$ $(x',l))$ where
                $x' \neq x$. That is, the refutation
                contains a counter-example
                of an instance $x'$ that has the same label $l$ as $x$
                but $(x',l)$ is not derivable by $p$
            \item The proof $p$ is overly-general.
                A possible refutation is then
                $(\mathit{overgeneral}(E)$ $because$ $(x',l'))$,
                where $x'$ is possibly different to $x$, and $l' \neq l$.
                That is, the refutation contains a counter-example 
                of an instance $x'$ with a label $l'$ that is distinct from $l$,
                but $(x',l)$ is incorrectly derivable by $p$.
        \end{itemize}
       
\end{description}
 
 \noindent
 
 \subsubsection{Practicalities}
 \label{sec:prac}
 
Constructing a workable collaborative system requires more than just the
bare-bones formalisation we have described. Let us look again at
Fig.~\ref{fig:covidx}. There the radiologist does indeed provide
a refutation of the machine's explanation. But in what we are now
proposing, the radiologist must also indicate the area of the right
lobe containing the opacification. This will constitute the counter-example
in the message sent to the machine. Identifying counter-examples are
also not the end of the story. The machine, on receiving a counter-example,
has to {\em {do}} something with it: either convince the human that their
refutation is invalid, or update its hypothesis. Nothing is more
guaranteed to lose the human's patience with the machine than if it makes
the same kind of mistakes on the same kind of data: unfortunately, this is exactly
what would result with once-off model construction such as the approach
used in \cite{covid}. 
However, we know from PEIRS that revison of tree-structured models is
feasible and desirable, as it allowed the machine not to
repeat its previous mistakes. Thus, it was not
simply the use of refutations, but the resulting corrections
that were responsible for developing a sustained collaboration between the
human and machine (PEIRS was in routine clinical use for about 4 years and had
accumulated about 2000 rules, each resulting from feedback from a pathologist).

There are some important differences between PEIRS and 
what is being proposed here. First, 
PEIRS did not have any independent automated ability
for hypothesis construction.
Instead, the decision-maker was
involved in both identifying refutations and guiding
updates to the machine's hypothesis. The role of the
machine was primarily to focus
the decision-maker's attention to potentially problematic aspects in
the data (in itself no mean feat, since the data
could often be over 50 separate time-series trajectories,
many with gaps, and spanning time-periods of a few hours to several days).
Thus, in principle, the machine's hypothesis could not infer
anything different to the human decision-maker, removing
the possibility of suggesting anything new on data for which
the human decision-maker was uncertain. Thus,
it is not possible for the human-performance axiom
to hold in principle (in practice, the machine can
help maintain human performance by helping manage fatigue).
A second difference to what is proposed here is that
the domain-knowledge provided by the decision maker was
restricted in PEIRS to a one-off identification
of a vocabulary of relevant features and functions.
Some subsequent construction of
new features is possible, but only using
functions and features defined over the original vocabulary.
This does restrict the expressive power of hypotheses that can be
constructed by
the machine. The human-refutability
axiom will not apply, for example, if instances with different labels
become indistinguishable because of the restricted
vocabulary. Since the human will not be able to construct a new rule,
no corresponding revision of the machine's hypothesis will occur and
the machine-performance axiom will also not hold.
In contrast, what we are suggesting allows updates of
the domain-knowledge provided by the decision-maker to the machine.
This includes the vocabulary. Additionally, modern-day ML systems like
deep neural networks that routinely construct internal representations that extend
the initial vocabulary: a significant challenge arises in communicating
these new concepts in an intelligible way to the human.\footnote{
By this we mean an explanation that is at least refutable in principle. One
direction is that of proxy explanations using symbolic models that are consistent
with the neural models, at least to some limited local extent \cite{Ribe:etal:p:2016}. The other
area of interest is that of {\em self-explaining\/} deep networks, in which each internal
construct has a clear meaning \cite{selfexplain}.} What will probably survive from the
PEIRS approach though is basic principle of repeated iterations of conjectures and refutations. Even there, a re-examination of the
principle will be needed, since it assumes a human is always
capable of correctly refuting a machine's conjecture. There is 
experimental evidence of how human plus machine performance
may be impacted, for example, due to ``automation bias'' where human specialists can fail to correctly override machine errors~\cite{Lyel:etal:j:2017}, or due to a lack of intelligibility of the machine's explanations~\cite{mugg:expl}.

 At this point, the reader may be wondering
 whether the collaborative process we envisage terminates at all, and if
 it does, in what kind of state?  The question is really one of
 convergence in a mathematical sense. We
think this can be addressed mathematically in at least one of two ways: computational
learning theory, in the sense of arriving at an acceptable approximation
to the oracle after some (bounded number of) queries; or the theory
of co-operative games in which some players have primacy over others
(for example, the oracle over the human, and the human over the
machine). But there is no immediate urgency regarding the definition of convergence.
For the present, we can assume 
 that after some exchange of messages the human or machine or both
 will end up updating their domain-knowledge and corresponding hypothesis.
 We can therefore imagine the collaboration to be
 a described by sequence of hypotheses
 $(H_0,H_1,\ldots,H_k,\ldots)$ for human
 and $(M_0,M_1,\ldots,M_k,\ldots)$ for machine; such that, for each
 point $i > 0$, either $M_i \neq M_{i-1}$ or $H_i \neq H_{i-1}$.
More usefully,
  if we can ensure on each iteration $i > 0$, either one of
  the following occurs:
 
    \begin{description}
        \item[Machine-revision.] There is a transfer of domain-knowledge from
            the human--including refutations--which results in subsequent improvement in machine-performance; or
        \item[Human-revision.] There is a transfer of recommendations from the
            machine, at least one of whose explanations is not refuted, and which
            results in subsequent improvement in human-performance.
    \end{description}
 
 \noindent
 then we are assured that at least one of the intelligibility axioms
 holds on each iteration.
 
How much of what we have described so far be achieved now? At least within
the world of symbolic ML, there
is a long history of work that is
explicitly concerned with human-machine interaction.
This includes a very early recognition of the
role of domain-knowledge in hypothesis construction \cite{plotkin:thesis},
Michie's characterisation of ultra-strong machine
learning \cite{dm:ml}, down to recent studies
with human-subjects on identifying a ``cognitive window''
relating symbolic descriptions constructed by ML to human
performance \cite{mugg:expl}. Implementations have
similarly included mechanisms for generating examples
and questions for a human  \cite{sammut:marvin}, refutation-driven
revision of hypotheses \cite{PEIRS:j:1993,More:Crop:x:2021} down to
updates allowing a re-shaping of the
hypothesis space, based on using and extending the domain-knowledge
provided \cite{Crop:More:j:2021}.

Despite this substantial body of  work--now spanning over
4 decades--on aspects of human-to-machine and machine-to-human
information sharing,  there is in fact surprisingly little that
has emerged in the form of a deployed tool that conducts a sustained human-machine collaboration (PEIRS being a notable exception,
albeit with strict limitations on expressivity and ML capabilities).
The actual use of symbolic ML has thus not exploited its
ability to enable intelligible human-machine collaboration.
But part-successes  suggest that, with symbolic ML
at least, such a collaboration is feasible with
current technology.\footnote{It
is unsurprising, given this apparent suitability of symbolic ML and
our own involvement in the area, that the basic formalisation we provided
earlier has strong logical overtones, based on an existing
implementation of incremental
learning designed for collaborative ML \cite{aleph}. A
recent example of collaborative use of an extended
version of this tool is in \cite{moyle}. }
    No doubt this will also extend to the use of newer forms of ML that
    combine neural- and symbolic-representations, which can draw on the work done in
    the purely symbolic realm (evidence of this already
    emerging in the design of explanation-driven
    collaborative decision-making systems for medical images
    \cite{Schm:Finz:p:2020}).
    The extent to which purely neural
    approaches to ML can be developed for intelligible collaboration
    with humans remains an open question.
 
 \section{Conclusion}
\label{sec:concl}
In this paper we have sought to look beyond the
current practice in Explainable Machine Learning. To
adopt a popular ML adjective, most of the current focus is on
``one-shot explainability''. That is, the machine produces
an explanation, without much regard to
whom the explanation is for, and even less concern about what
it should do if the explanation is thought to have a problem.
The situation is somewhat akin to a car that
shows you are travelling at 212,085 furlongs per fortnight,
and then does nothing if you apply the brakes. Confidence
in the car will naturally be dented.

It is our
contention that designers of ML-based decision-support tools
cannot afford this kind of solipsism. Instead, by
design the tools must be concerned with mutual intelligibility.
Properties that will assist 2-way intelligibility
are:

\begin{itemize}
    \item We would want the machine to have access
        to all information the human decision-maker thinks is relevant
        in a form that the machine can use; and
    \item We would want human and machine to provide explanations
        in a form that the other can inspect and
        evaluate critically, and for each to respond meaningfully
        to that evaluation.
\end{itemize}

Both properties are easier to state than to achieve. Let us
consider each in turn. For the first,
the human decision-maker may not have a clear idea of knowledge that is relevant for the machine.
In practice, we have to live with what could be {\em potentially\/}
relevant, but then we have another problem: how is this information to be
communicated to the machine?
McCarthy \cite{McCa:p:1959} envisaged this would
be done as statements in a formal language
that would be directly manipulated by the machine.
But this does not have to be necessarily the case.
Instead, all that may be needed is for domain-knowledge to
be translatable into a form that transforms some
or all of the inputs of the ML system, for
example: the data, utility function, structure or
parameters in Fig.~\ref{fig:mlsys}. However, in
the short- to medium-term, we believe this will
require the human decision-maker to have a working knowledge
of the ML system, or to employ someone who does.
In the longer term, neither of these options are
practical, and decision-support tools will
need mechanisms to receive information in natural language,
and perform any manipulations
to its inputs internally. Some progress is being made on
processing information in a natural language (see for example,
\cite{clark:nlproofs}), but we are still far from being able to do this well. 
Assuming we have the human-supplied information in some
machine-usable form, the machine may still need
to ask the decision-maker questions to clarify any ambiguities
or inconsistencies. To resolve these would undoubtedly
need the machine
to be able to generate its own data, and ask questions about them.
How should it communicate these questions and receive answers? Again,
in the long-run, natural human-computer interaction techniques~\cite{DAmi:etal:c:2010} seem inevitable. 

In order for the human to evaluate critically the machine's advice, the advice
will have to employ concepts that the decision-maker can recognise.
In the near-term, the machine can achieve this in one of two ways. First,
the machine can elect to employ only those concepts that are already known to the
decision-maker. The identification of Covid patients in the previous
section is an example: the features extracted from chest X-rays were
restricted to those identified by a radiologist. A second way is for
the machine to show instantiations through some textual
or visual means. For example, ML systems that attempted to discover chess concepts showed board positions exemplifying the concept.
The chess-expert may then be able to map the machine-identified
concept to some part of their chess vocabulary
(such as ``this is really about Kings in opposition''). In the
long-term, as the problems, data and ML systems get more complex,
it will not be possible to engineer an adequate set of features
beforehand, and mapping to known concepts would not be immediate.
We expect this will entail a kind of dialectical exchange of
questions and answers between the human and the machine,
as the human decision-maker attempts to understand the why the
machine is proposing what it does. So intelligibility
of the machine's explanation would be reached, but not
without effort. Intelligibility of human explanations
could prove even more difficult, especially if those
explanations are in a natural language. This will may
well require the machine to contain language models for
approximating idealised forms of logical reasoning and
contradiction detection, which can then form the basis
of the machine's refutation.

But while the issues are challenging, the problem is not insoluble.
We have shown examples
from the literature where attempts at
intelligibility have yielded significant benefits.
As decisions and data get more complex, it is evident that if
an ML-based decision-support tool is to be of on-going value to the
decision-maker, then we must design for both machine and human to
learn from each other's recommendations. Our
3Rs for embarking on the design of 2-way intelligible decision-support
with ML are therefore:
$Refute$, $Revise$, $Repeat$.

\vspace*{0.5cm}
\noindent
{\scriptsize{
{\bf {Acknowledgements.}} AS is a Visiting Professor at Macquarie University, Sydney
and a Visiting Professorial Fellow at UNSW, Sydney. He is also the Class of 1981
Chair Professor at BITS Pilani, Goa,
the Head of the Anuradha and Prashant Palakurthi Centre for AI Research (APPCAIR) at BITS Pilani, and a Research Associate
at TCS Research. Many of the
results reported here are from collaborative work done by the authors with
colleagues at several institutions. We would especially like to acknowledge the
role played by: Tirtharaj Dash, Rishabh Khincha, Soundarya Khincha, Lovekesh
Vig, Arijit Roy, Gautam Shroff, Paul Compton, Ross King and Stephen Muggleton. AS and MB owe a debt of gratitude to Donald Michie, who shaped much of their
thinking on the views expressed in this paper.
}}

\end{document}